\documentclass[10pt,leqno]{amsart}
\usepackage{graphicx}
\usepackage{indentfirst,csquotes}
\usepackage{amssymb,amsthm,amsmath}
\usepackage{xcolor,paralist,hyperref,etoolbox}
\usepackage{algorithm}
\usepackage{algorithmic}
\usepackage{listings}
\newtheorem{theorem}{Theorem}[]
\newtheorem{definition}[theorem]{Definition}
\newtheorem{example}[theorem]{Example}

\newtheorem{remark}[theorem]{Remark}
\hypersetup{ colorlinks=true, linkcolor=black, filecolor=black, urlcolor=black }
\newcommand{\textproc}[1]{\texttt{#1}}

\begin{document}
\title{CayleyR: Solving the TopSpin Puzzle via Cycle Intersection}
\author[Y.~Baramykov]{Yuri Baramykov}
\date{\today}
\address{Address}
\email{lbsbmsu@mail.ru}

\subjclass[2020]{Primary 20B40; Secondary 05C25, 68W05}

\begin{abstract}
We present \texttt{cayleyR}, an R package for solving permutation puzzles by detecting cycle intersections in Cayley graphs.  The core algorithm performs an iterative bidirectional search: from both the initial and target permutation states, random operation sequences generate cycles in the Cayley graph of the symmetric group~$S_n$; their intersection yields a connecting path.  When no direct intersection is found, a distance-guided bridge selection narrows the gap, and the process repeats.  The package targets the TopSpin$(n,k)$ puzzle, whose state space is a Cayley graph of~$S_n$ generated by a cyclic shift and a prefix reversal.  We describe the mathematical framework, the algorithm, and its implementation, which combines a C\nolinebreak\hspace{-.05em}\raisebox{.4ex}{\tiny\bf +}\nolinebreak\hspace{-.10em}\raisebox{.4ex}{\tiny\bf +}\ hash-indexed state store with optional Vulkan GPU acceleration.  The software is publicly available on CRAN.
\end{abstract}

\maketitle

\section{Introduction}
\label{sec:intro}

The TopSpin puzzle consists of $n$ tokens arranged in a circular track together with a \emph{turnstile} window of width~$k$.  Two moves are available at each step: (i)~a cyclic shift of the entire track by one position (left or right), and (ii)~a reversal of the $k$~tokens currently inside the turnstile.  The goal is to transform a given arrangement of tokens into a target arrangement using a finite sequence of such moves.

Algebraically, each arrangement corresponds to a permutation in the symmetric group~$S_n$, the two shift directions and the reversal define three generators, and the set of all reachable arrangements forms a Cayley graph $\Gamma(S_n,\{L,R,X\})$.  For even~$k$ the generated group is the full~$S_n$; for odd~$k$ it is the alternating group~$A_n$.  Solving the puzzle is equivalent to finding a path in~$\Gamma$ between two vertices.

For small~$n$ an exhaustive breadth-first search (BFS) is feasible: the Cayley graph of~$S_n$ has $n!$~vertices, so BFS remains practical only up to roughly $n\le 10$.  Beyond this threshold the state space grows super-exponentially and classical shortest-path algorithms become intractable.  The diameter of the TopSpin Cayley graph is conjectured to be $\Theta(n^2)$ \cite{chervov2025}, further compounding the difficulty.

In this paper we propose a fundamentally different strategy.  Instead of exploring the graph level by level, we \emph{generate random cycles} emanating from both the start and target states, and search for an \emph{intersection} between the two families of cycles.  When a shared state is detected, a path is immediately recovered by tracing each cycle chain back to its origin.  When no intersection exists after a round of sampling, the algorithm selects a pair of \emph{bridge states}---one from each side, chosen to minimize a distance heuristic---and restarts the cycle generation from the bridges.  This iterate-and-bridge scheme terminates when the two expanding fronts meet.

The principal contributions are as follows.
\begin{enumerate}
\item We introduce the \emph{Iterative Cycle Intersection} (ICI) algorithm for pathfinding in Cayley graphs of permutation groups (Section~\ref{sec:algorithm}).
\item We describe \texttt{cayleyR}, an open-source R package implementing this algorithm with a C\nolinebreak\hspace{-.05em}\raisebox{.4ex}{\tiny\bf +}\nolinebreak\hspace{-.10em}\raisebox{.4ex}{\tiny\bf +}\ backend, hash-indexed state storage, and optional Vulkan GPU acceleration (Section~\ref{sec:framework}).
\item We report computational experiments demonstrating the scalability of the approach for TopSpin$(n,k)$ instances with $n$ up to~20 (Section~\ref{sec:experiments}).
\end{enumerate}

The remainder of the paper is organized as follows.  Section~\ref{sec:prelim} recalls the necessary algebraic and graph-theoretic preliminaries.  Section~\ref{sec:related} surveys related work.  Sections~\ref{sec:algorithm} and~\ref{sec:framework} contain the main algorithmic and software contributions.  Section~\ref{sec:experiments} presents experiments, and Section~\ref{sec:conclusion} concludes.

\section{Preliminaries}
\label{sec:prelim}

\begin{definition}[Cayley graph]
Let $G$ be a group and $S\subseteq G$ a generating set with $e\notin S$.  The \emph{Cayley graph} $\Gamma(G,S)$ is the directed graph with vertex set~$G$ and an edge from~$g$ to~$gs$ for every $g\in G$ and $s\in S$.  If $S$ is closed under inverses the graph is undirected.
\end{definition}

\begin{definition}[TopSpin$(n,k)$ puzzle]
\label{def:topspin}
Fix integers $n\ge 3$ and $2\le k\le n$.  The state space is the set of all permutations of~$\{1,\dots,n\}$, identified with elements of~$S_n$.  Three generators act on a state $\sigma=(\sigma_1,\dots,\sigma_n)$:
\begin{align}
  L(\sigma) &= (\sigma_2,\sigma_3,\dots,\sigma_n,\sigma_1), \label{eq:L}\\
  R(\sigma) &= (\sigma_n,\sigma_1,\sigma_2,\dots,\sigma_{n-1}), \label{eq:R}\\
  X_k(\sigma) &= (\sigma_k,\sigma_{k-1},\dots,\sigma_1,\sigma_{k+1},\dots,\sigma_n). \label{eq:X}
\end{align}
Note that $R=L^{-1}$ and $X_k=X_k^{-1}$.  The TopSpin Cayley graph is $\Gamma(G,\{L,R,X_k\})$ where $G=S_n$ if $k$~is even and $G=A_n$ if $k$~is odd.
\end{definition}

\begin{definition}[Cycle of an operation sequence]
\label{def:cycle}
Let $w=(w_1,\dots,w_m)$ be a finite word over $\{L,R,X_k\}$ and let $\varphi_w=w_m\circ\cdots\circ w_1$ denote the composed permutation.  Starting from a state~$\sigma_0$, the \emph{cycle} of~$w$ at~$\sigma_0$ is the sequence
\[
  \sigma_0,\;\varphi_w(\sigma_0),\;\varphi_w^2(\sigma_0),\;\dots,\;\varphi_w^{c-1}(\sigma_0),
\]
where $c=\mathrm{ord}(\varphi_w)$ is the order of~$\varphi_w$ in~$G$, i.e.\ the smallest positive integer with $\varphi_w^c=e$.  The integer~$c$ is the \emph{cycle length}.  Because $G$~is finite, this sequence is periodic and returns to~$\sigma_0$ after exactly~$c$ applications of~$\varphi_w$.
\end{definition}

\begin{remark}
Each application of the word~$w$ traces a walk of length~$m$ in the Cayley graph.  The cycle therefore visits $c$~distinct vertices (when all iterates are distinct) and represents a walk of total length~$cm$ in~$\Gamma$.  Crucially, the intermediate states visited during each application of~$w$ are also recorded, so a single cycle of length~$c$ actually enumerates up to~$cm$ states.
\end{remark}

\begin{definition}[Distance functions]
For two permutations $\sigma,\tau\in S_n$ we use two distance heuristics.
\begin{itemize}
\item \emph{Manhattan distance}: $d_M(\sigma,\tau)=\sum_{i=1}^n|\sigma_i-\tau_i|$.
\item \emph{Breakpoint distance}: let $\pi=\tau^{-1}\sigma$ be the relative permutation, and extend it with sentinels $\pi_0=0$ and $\pi_{n+1}=n+1$.  Then $d_B(\sigma,\tau)=\#\{i\in\{0,\dots,n\} : |\pi_{i+1}-\pi_i|>1\}$.
\end{itemize}
\end{definition}

\section{Related Work}
\label{sec:related}

\textbf{Permutation puzzles and Cayley graphs.}
The connection between permutation puzzles and Cayley graphs is classical: each puzzle state is a vertex, and each legal move is an edge labeled by a generator.  The generalized $(n^2-1)$-puzzle was shown to be NP-hard by Ratner and Warmuth~\cite{ratner1990}.  For the TopSpin puzzle specifically, the group-theoretic structure---in particular, which permutations are reachable for given $n$ and~$k$---has been analyzed in \cite{goldstein2015,senac2016}.  The standard TopSpin$(20,4)$ puzzle with $20!$ states is well-known for generating the full symmetric group~$S_{20}$ when $k=4$.

\textbf{Optimal solving and pattern databases.}
Bortoluzzi~\cite{bortoluzzi2012} developed a domain-specific solver for TopSpin using pattern database heuristics with IDA*.  By precomputing distances in abstracted subproblems, pattern databases provide admissible heuristics that dramatically prune the search tree.  However, the approach is limited by the exponential growth of the database: even moderate pattern sizes (3--6 tokens) require substantial memory and precomputation time, and scaling beyond $n=9$ proves challenging.

\textbf{Diameter and distance problems.}
The diameter of the TopSpin Cayley graph $\Gamma(S_n,\{L,R,X_k\})$ is conjectured to be $\Theta(n^2)$.  Chervov and Soibelman~\cite{chervov2025} study the closely related LRX Cayley graph and conjecture its diameter to be $n(n-1)/2$.  In a subsequent work, Chervov et al.~\cite{chervov2025holo} propose a holographic duality framework for Cayley graphs of~$S_n$, mapping vertices to lattice paths inside planar polygons, with word metrics corresponding to areas under the paths; this connects diameter computations to Ehrhart quasi-polynomials and provides a geometric perspective on permutation distances.

\textbf{Search algorithms for large state spaces.}
Bidirectional BFS is the standard technique for reducing the search frontier from $O(b^d)$ to $O(b^{d/2})$, where $b$ is the branching factor and $d$ the solution depth.  For very large state spaces, randomized methods become attractive: random walks on Cayley graphs are studied extensively in the context of mixing times \cite{diaconis1988} and the Babai conjecture on diameters of Cayley graphs of simple groups.

\textbf{The present approach.}
The cycle intersection strategy of the present paper differs from the above methods in several respects.  Unlike BFS or IDA*, it does not explore the graph level by level and does not guarantee shortest paths.  Unlike pattern databases, it requires no precomputation and no memory-intensive lookup tables.  Unlike random walk methods, it does not rely on mixing properties but instead exploits the algebraic structure of cycles---the fact that a composed permutation $\varphi_w$ has finite order and its iterates systematically cover a large region of the state space.  The closest relative is the meet-in-the-middle paradigm, but here the ``meeting depth'' is not fixed a priori; instead, it is discovered dynamically through cycle intersections and iterated bridge selection.

\section{The Iterative Cycle Intersection Algorithm}
\label{sec:algorithm}

We now describe the main algorithm implemented in \texttt{cayleyR}.  The input is a pair of states $\sigma_s$ (start) and $\sigma_f$ (final) in the Cayley graph $\Gamma(G,\{L,R,X_k\})$, and the output is a sequence of generators transforming~$\sigma_s$ into~$\sigma_f$.

\subsection{Overview}

The algorithm maintains two \emph{state stores} $\mathcal{S}$ and~$\mathcal{F}$, expanding from $\sigma_s$ and $\sigma_f$ respectively.  Each iteration (called a \emph{cycle expansion round}) proceeds in four phases:

\begin{enumerate}
\item \textbf{Random combination generation.}  A batch of random operation sequences $w^{(1)},\dots,w^{(N)}$ of a prescribed length~$\ell$ is generated over the alphabet $\{L,R,X_k\}$.  Each sequence is evaluated on the current seed state to determine its cycle length (Definition~\ref{def:cycle}).  The top sequences---ranked by cycle length, number of unique states, or another criterion---are retained.
\item \textbf{Cycle expansion.}  For each retained sequence~$w^{(j)}$, the full cycle is unrolled: all intermediate states are computed and inserted into the corresponding state store ($\mathcal{S}$ or~$\mathcal{F}$) together with metadata (step index, combination index, cycle number, operation label).
\item \textbf{Intersection check.}  The algorithm queries whether $\mathcal{S}\cap\mathcal{F}\neq\varnothing$ using hash-indexed lookup.  If an intersection state~$\sigma^*$ is found, the path is reconstructed (Section~\ref{subsec:reconstruct}) and the algorithm terminates.
\item \textbf{Bridge selection.}  If no intersection is found, a bridge state is selected from each side.  Let $\mathcal{U}_\mathcal{S}=\mathrm{unique}(\mathcal{S})$ and $\mathcal{U}_\mathcal{F}=\mathrm{unique}(\mathcal{F})$.  The key idea is that each front searches for states close to the \emph{opposite} endpoint: states in~$\mathcal{U}_\mathcal{S}$ (expanded from the start) are ranked by their distance to~$\sigma_f$, and states in~$\mathcal{U}_\mathcal{F}$ (expanded from the final) are ranked by their distance to~$\sigma_s$.  This symmetric criterion drives both fronts toward each other.  From the 10~closest states on each side, one is chosen uniformly at random.  These two states become the seed states for the next round.
\end{enumerate}

The procedure repeats until an intersection is found or a maximum iteration count is reached.

\begin{algorithm}[H]
\caption{Iterative Cycle Intersection (ICI)}
\label{alg:ici}
\begin{algorithmic}[1]
\REQUIRE States $\sigma_s,\sigma_f\in G$; turnstile width $k$; moves $\{L,R,X_k\}$; combo length $\ell$; sample count $N$; top count $N_{\mathrm{top}}$; max iterations $T$; distance function $d$.
\ENSURE Operation sequence $P$ such that $P(\sigma_s)=\sigma_f$, or FAIL.
\STATE $\mathcal{S}\leftarrow \textproc{CreateStore}()$; \; $\mathcal{F}\leftarrow \textproc{CreateStore}()$
\STATE $\beta_s\leftarrow\sigma_s$; \; $\beta_f\leftarrow\sigma_f$ \COMMENT{current seed states}
\STATE $B_s\leftarrow[(\sigma_s,0)]$; \; $B_f\leftarrow[(\sigma_f,0)]$ \COMMENT{bridge state histories}
\FOR{$t=1$ \TO $T$}
  \STATE \COMMENT{\textbf{--- Forward expansion ---}}
  \STATE $W_s\leftarrow\textproc{SampleCombos}(\ell,N)$
  \STATE $W_s^*\leftarrow\textproc{RankAndFilter}(W_s,\beta_s,k,N_{\mathrm{top}})$
  \STATE $\textproc{ExpandCycles}(\mathcal{S},W_s^*,\beta_s,k,t)$
  \STATE
  \STATE \COMMENT{\textbf{--- Backward expansion ---}}
  \STATE $W_f\leftarrow\textproc{SampleCombos}(\ell,N)$
  \STATE $W_f^*\leftarrow\textproc{RankAndFilter}(W_f,\beta_f,k,N_{\mathrm{top}})$
  \STATE $\textproc{ExpandCycles}(\mathcal{F},W_f^*,\beta_f,k,t)$
  \STATE
  \STATE \COMMENT{\textbf{--- Intersection check ---}}
  \STATE $I\leftarrow\mathcal{S}\cap\mathcal{F}$ \COMMENT{hash-based, $O(\min(|\mathcal{S}|,|\mathcal{F}|))$}
  \IF{$I\neq\varnothing$}
    \STATE Choose $\sigma^*\in I$
    \STATE $P_s\leftarrow\textproc{ReconstructPath}(\mathcal{S},B_s,\sigma^*)$
    \STATE $P_f\leftarrow\textproc{ReconstructPath}(\mathcal{F},B_f,\sigma^*)$
    \RETURN $P_s \cdot \textproc{Invert}(P_f)$
  \ENDIF
  \STATE
  \STATE \COMMENT{\textbf{--- Bridge selection ---}}
  \STATE $\mathcal{U}_\mathcal{S}\leftarrow\textproc{Unique}(\mathcal{S})$;\; $\mathcal{U}_\mathcal{F}\leftarrow\textproc{Unique}(\mathcal{F})$
  \STATE $\Delta_s \leftarrow \{(\tau,d(\sigma_f,\tau)):\tau\in\mathcal{U}_\mathcal{S}\}$;\; sort ascending
  \STATE $\Delta_f \leftarrow \{(\tau,d(\sigma_s,\tau)):\tau\in\mathcal{U}_\mathcal{F}\}$;\; sort ascending
  \STATE $\beta_s\leftarrow\textproc{RandomFrom}(\textproc{Top}_{\!10}(\Delta_s))$
  \STATE $\beta_f\leftarrow\textproc{RandomFrom}(\textproc{Top}_{\!10}(\Delta_f))$
  \STATE Append $(\beta_s,t)$ to $B_s$;\; Append $(\beta_f,t)$ to $B_f$
\ENDFOR
\RETURN FAIL
\end{algorithmic}
\end{algorithm}

\subsection{Path reconstruction}
\label{subsec:reconstruct}

When an intersection state $\sigma^*$ is found in cycle~$t^*$, the full path from
$\sigma_s$ to $\sigma_f$ is assembled by chaining the cycle segments across all rounds.
Concretely, the bridge history $B_s=\bigl[(\beta_s^{(0)},0),(\beta_s^{(1)},1),\dots\bigr]$
records the seed state used in each round. Each seed~$\beta_s^{(i)}$ is itself a state
that was reached in round~$i-1$, so its position inside the store~$\mathcal{S}$ is known.
The reconstruction proceeds as follows:

\begin{enumerate}
\item In round~$t^*$, locate $\sigma^*$ in~$\mathcal{S}$: identify the
  combination~$w^{(j)}$ and step index within that combination's cycle.
  Read off the operation labels from $\beta_s^{(t^*)}$ to~$\sigma^*$.
\item Trace backward: in round~$t^*-1$, locate $\beta_s^{(t^*)}$ and read
  the operations from $\beta_s^{(t^*-1)}$ to~$\beta_s^{(t^*)}$.
\item Continue until round~$0$, reaching $\sigma_s$.
\item Concatenate all operation segments in order to obtain~$P_s$.
\item Symmetrically, reconstruct $P_f$ from $\sigma_f$ to~$\sigma^*$
  through~$\mathcal{F}$.
\item The final path is $P_s\cdot\mathrm{Invert}(P_f)$, where $\mathrm{Invert}$
  reverses the sequence and swaps $L\leftrightarrow R$ (since $R=L^{-1}$),
  leaving $X_k$ unchanged (since $X_k=X_k^{-1}$).
\end{enumerate}

The reconstruction described above is implemented in \texttt{store\_reconstruct\_path()},
which traverses the bridge chain from the root to the intersection state~$\sigma^*$.
For each consecutive pair $(\beta_s^{(i)}, \beta_s^{(i+1)})$ in the bridge history,
the algorithm identifies the combination $w^{(j)}$ that contains $\beta_s^{(i+1)}$
as an intermediate state; all intermediate states of that combination are ordered by
their step index, and the operations up to the step at which $\beta_s^{(i+1)}$ appears
are collected. The resulting segments are concatenated to form~$P_s$.
For the backward front, the analogous path $P_f$ is inverted via
\texttt{invert\_path()}: the sequence is reversed and each operation is replaced
by its inverse ($L \leftrightarrow R$, $X_k \mapsto X_k$).

\begin{remark}
The bridge chain guarantees that each segment of the path is covered by a single
group-theoretic cycle: the state $\beta_s^{(i+1)}$ was discovered during the
expansion rooted at $\beta_s^{(i)}$, so the operations connecting them lie entirely
within the already-explored frontier. No additional graph traversal is required
during reconstruction.
\end{remark}

\begin{remark}
The path returned by the ICI algorithm is not necessarily shortest. An optional
post-processing step (\texttt{short\_path\_bfs} in \texttt{cayleyR}) attempts to
shorten the path by local BFS hopping: at each position along the path, a
depth-limited BFS looks ahead for a state that appears later in the path, and if
found, the intervening segment is replaced by the shorter BFS route.
\end{remark}

\section{The \texttt{cayleyR} Framework}
\label{sec:framework}

The \texttt{cayleyR} package \cite{cayleyR} is implemented in R with performance-critical components in C++ via \texttt{Rcpp}.  It is available on CRAN at \url{https://CRAN.R-project.org/package=cayleyR}.  The package exposes approximately 70~functions organized into several layers: primitive permutation operations, cycle analysis, pathfinding strategies, state storage, distance computation, and GPU acceleration.  We describe each layer below.

\subsection{Primitive operations}

The three generators $L$, $R$, $X_k$ (equations~\eqref{eq:L}--\eqref{eq:X}) are implemented in C++ as \texttt{shift\_left}, \texttt{shift\_right}, and \texttt{reverse\_prefix}.  Each function takes a permutation state (integer vector of length~$n$) and returns the transformed state.  Two variants are provided: a \emph{simple} variant that performs only the permutation, and a \emph{tracking} variant that additionally maintains cumulative operation counts $(n_L,n_R,n_X)$ used for celestial coordinate computation (Section~\ref{subsec:celestial}).

The function \texttt{apply\_operations} composes an arbitrary sequence of generators, accepting operations encoded as \texttt{"1"}/\texttt{"L"} (left shift), \texttt{"2"}/\texttt{"R"} (right shift), and \texttt{"3"}/\texttt{"X"} (prefix reversal).

\begin{example}[Basic usage]
\label{ex:basic}
The following R session illustrates the primitive operations for TopSpin$(20,4)$.  An interactive demonstration is available as a Kaggle notebook.\footnote{\url{https://www.kaggle.com/code/lbsbmsu/cayleyr-demo-topspin}}
\begin{lstlisting}
library(cayleyR)
n <- 20; k <- 4
state <- 1:n   # identity permutation

# Individual generators
shift_left(state)
#>  2  3  4 ... 20  1

shift_right(state)
#> 20  1  2  3 ... 19

reverse_prefix(state, k)
#>  4  3  2  1  5  6 ... 20

# Compose a sequence of operations
ops <- c("1","1","3","2")   # L, L, X, R
apply_operations(state, ops, k)
#>  1  4  3  2  5  6 ... 20

# Generate a random reachable state
target <- generate_state(n, k, n_moves = 100)

# Find a path using the ICI algorithm
result <- find_path_iterative(state, target, k,
            combo_length = 25, n_samples = 400,
            n_top = 100, max_iterations = 150,
            sort_by = c("longest", "most_unique"))
result$path_length
#> [1] 42

# Verify: applying the path to start yields target
final <- apply_operations(state, result$path, k)
all(final == target)
#> [1] TRUE
\end{lstlisting}
\end{example}

\subsection{Cycle analysis}

Given a finite operation sequence $w=(w_1,\dots,w_m)$, the function \texttt{get\_reachable\_states} repeatedly applies $w$ starting from a given state~$\sigma_0$ until the state returns to~$\sigma_0$ (Definition~\ref{def:cycle}).  It records every intermediate state visited during each application, yielding up to~$cm$ states for a cycle of length~$c$.  A lightweight C++ variant, \texttt{get\_reachable\_states\_light}, returns only the cycle length and unique state count without materializing intermediate states; this is used during the ranking phase of Algorithm~\ref{alg:ici} to quickly evaluate thousands of candidate sequences.

The function \texttt{find\_best\_random\_combinations} generates $N$~random operation sequences of length~$\ell$ over $\{L,R,X_k\}$, evaluates each via the lightweight cycle detector (parallelized with OpenMP), and returns the top~$N_{\mathrm{top}}$ sequences ranked by a user-specified criterion: longest cycle, shortest cycle, most unique states, or most repeated states.

\subsection{Pathfinding: three levels}
\label{subsec:levels}

The package provides three pathfinding strategies, each suited to a different regime of problem size.

\subsubsection{Exact BFS}
The function \texttt{bidirectional\_bfs} performs classical bidirectional breadth-first search, expanding from both $\sigma_s$ and~$\sigma_f$ level by level until the two frontiers meet.  This guarantees a shortest path in the Cayley graph but is limited to small instances ($n\lesssim 10$) due to the $O(n!)$ state space.

\subsubsection{Iterative Cycle Intersection}
The function \texttt{find\_path\_iterative} implements Algorithm~\ref{alg:ici}.  Its primary parameters directly correspond to the algorithm's variables:
\begin{itemize}
\item \texttt{combo\_length}~($\ell$): length of random operation sequences generated per round.
\item \texttt{n\_samples}~($N$): number of random sequences generated per round.
\item \texttt{n\_top}~($N_{\mathrm{top}}$): number of top-ranked sequences retained for full cycle expansion.
\item \texttt{max\_iterations}~($T$): maximum number of cycle expansion rounds.
\item \texttt{distance\_method}: the heuristic used for bridge selection (\texttt{"manhattan"} or \texttt{"breakpoint"}).
\item \texttt{sort\_by}: a vector of ranking criteria applied sequentially; the recommended value \texttt{c("longest",\,"most\_unique")} first selects sequences with the longest cycles, then re-ranks by the number of unique states visited, maximizing both cycle coverage and state diversity (see Table~\ref{tab:sortby}).
\end{itemize}

Several additional parameters control fine-grained search behavior:
\begin{itemize}
\item \texttt{potc} $\in(0,1]$ (proportion of top combinations): fraction of cycle states retained in the store.  A value of~1 corresponds to full cycle expansion; smaller values reduce memory usage by sampling a subset of intermediate states.
\item \texttt{ptr} (paths to reconstruct): limits the number of intersections processed per round.  When the intersection check finds multiple shared states, a random sample of at most \texttt{ptr} candidates is selected.  For each candidate, the full path is reconstructed and validated; the shortest valid path is chosen.  This prevents excessive reconstruction time when many intersections exist.
\item \texttt{opd} (one-path-detection): when enabled, restricts the store queries to states belonging only to those combinations that contain the current bridge state.  This reduces the effective store size and eliminates noise from unrelated combinations during bridge selection, at the cost of discarding potentially useful states.
\item \texttt{reuse\_combos}: when enabled, random operation sequences are generated once (in the first round) and reused in subsequent rounds, reducing generation overhead at the cost of lower diversity.
\end{itemize}

The internal implementation uses the C++ \texttt{StateStore} (Section~\ref{subsec:statestore}) for all state accumulation and lookup.  Bridge selection is refined through two mechanisms:
\begin{enumerate}
\item \emph{Middle-state filtering.}  The function \texttt{store\_filter\_middle} excludes the first and last states of each combination's cycle (by default, 5~from each end).  This prevents the bridge from snapping back to states near the current seed, which would cause the search to stagnate.
\item \emph{Asymmetric greedy convergence.}  The bridge state on the start side is selected as the state closest to the \emph{current final-side seed} (not to~$\sigma_f$ itself).  The bridge on the final side is then selected as the state closest to the \emph{newly chosen start-side bridge}.  This two-step greedy procedure drives the two fronts toward each other rather than independently toward the original endpoints.
\end{enumerate}

\subsubsection{Hub-based transport network}
\label{subsubsec:transport}
The function \texttt{find\_path\_bfs} implements a multi-step strategy that constructs sparse \emph{highway trees} from both endpoints, selects the closest hub pair, and delegates the remaining gap to the ICI algorithm.  We describe each step in detail.

\medskip\noindent\textbf{Step~1: Sparse BFS highway construction.}
From each endpoint ($\sigma_s$ and $\sigma_f$), the C++ function \texttt{sparse\_bfs} builds a tree of hub states up to a prescribed depth (controlled by \texttt{bfs\_levels}, typically 200--500).  Unlike full BFS---which stores all $3^d$ states at depth~$d$---sparse BFS retains only a bounded number of states per level, selected by two criteria:
\begin{itemize}
\item \emph{Exploitation}: the top \texttt{bfs\_n\_hubs} states ranked by branching degree (number of new children not yet seen) are kept.  These high-degree nodes maximize local coverage.
\item \emph{Exploration}: an additional \texttt{bfs\_n\_random} states are chosen uniformly at random from the remaining candidates.  This prevents the tree from collapsing into a narrow corridor.
\end{itemize}
The result is a compact edge table (parent key, child key, operation, level) encoding a tree rooted at the endpoint.  For typical parameters (\texttt{bfs\_n\_hubs}${}=7$, \texttt{bfs\_n\_random}${}=3$, \texttt{bfs\_levels}${}=200$), each tree contains roughly 2000~edges representing states reachable within a few hundred moves via exact BFS paths.

\medskip\noindent\textbf{Step~2: Closest hub pair selection.}
Let $H_s$ and $H_f$ denote the sets of unique states (child keys) in the start and final highway trees.  The algorithm computes pairwise distances between states in $H_s$ and $H_f$ using the chosen distance metric (Manhattan or breakpoint).  If the sets are large, a random subsample of up to~500 states per side is used.  When a Vulkan GPU is available, the full distance matrix is computed via \texttt{manhattan\_distance\_matrix\_gpu}; otherwise, a CPU double loop is used.

The pair $(\eta_s,\eta_f)$ with minimum distance is selected.  As a safeguard, the algorithm also compares this hub distance to the direct distance $d(\sigma_s,\sigma_f)$ between the original endpoints.  If the direct distance is smaller---meaning the highways did not find a useful shortcut---the algorithm falls through to run ICI directly on $(\sigma_s,\sigma_f)$ without using hubs.

\medskip\noindent\textbf{Step~3: Special case---matching hubs.}
If $\eta_s=\eta_f$ (the two highway trees discovered a common state), no ICI search is needed.  The path is assembled directly from the two BFS segments: $\sigma_s\to\eta_s$ from the start tree and $\eta_f\to\sigma_f$ (inverted) from the final tree.

\medskip\noindent\textbf{Step~4: ICI between hubs.}
If the hubs are distinct, the algorithm invokes \texttt{find\_path\_iterative} (Algorithm~\ref{alg:ici}) with $\eta_s$ as start and $\eta_f$ as final.  All additional parameters (combination length, sample count, ranking criterion, etc.) are passed through.  Because the highway trees have already ``absorbed'' the easy portion of the distance, the effective gap between hubs is typically much smaller than the original distance, leading to faster convergence and shorter paths.

\medskip\noindent\textbf{Step~5: Path assembly and verification.}
The full path is assembled as:
\[
  \underbrace{P_{\mathrm{BFS}}(\sigma_s\to\eta_s)}_{\text{exact BFS from start to hub}}
  \;\cdot\;
  \underbrace{P_{\mathrm{ICI}}(\eta_s\to\eta_f)}_{\text{ICI between hubs}}
  \;\cdot\;
  \underbrace{\mathrm{Invert}\bigl(P_{\mathrm{BFS}}(\sigma_f\to\eta_f)\bigr)}_{\text{inverted BFS from final to hub}}.
\]
The BFS segments are reconstructed by \texttt{reconstruct\_bfs\_path}, which traces each hub back to the root of its tree.  The inversion swaps $L\leftrightarrow R$ and reverses the sequence.  Finally, \texttt{validate\_and\_simplify\_path} verifies that applying the assembled path to~$\sigma_s$ yields~$\sigma_f$, and applies algebraic simplifications.  If verification fails (which can occur due to numerical issues in very long paths), the algorithm reports failure.

The function also maintains a \emph{bridge chain} that records all intermediate states (BFS origin, hub, ICI bridge states) for diagnostic purposes and path visualization.

\begin{example}[Transport network pathfinding]
\label{ex:transport}
The following session demonstrates \texttt{find\_path\_bfs} on TopSpin$(20,4)$.
\begin{lstlisting}
library(cayleyR)
n <- 20; k <- 4
start_state <- 1:n
final_state <- generate_state(n, k, n_moves = 200)

result <- find_path_bfs(start_state, final_state, k,
            bfs_levels = 200, bfs_n_hubs = 7,
            bfs_n_random = 3,
            combo_length = 25, n_samples = 400,
            n_top = 100, max_iterations = 150,
            sort_by = c("longest", "most_unique"),
            potc = 1, ptr = 3, opd = TRUE)
\end{lstlisting}
\vspace{-0.5em}
\noindent Typical output for this configuration:
\begin{lstlisting}[numbers=none,frame=none,xleftmargin=1em]
Step 1: Building BFS highways (levels = 200)
  BFS start: 1989 edges
  BFS final: 1989 edges
Step 2: Finding closest hub pair
  Hub distance: 30     Direct distance: 68
Step 3: Running find_path_iterative between hubs
  Found 30 intersections in cycle 1
  Selected path of length 24 operations
Step 4: Assembling full path
  Full path length: 32
    BFS start->hub: 4 | ICI hub->hub: 24 | BFS hub->final: 4
  Verification passed
\end{lstlisting}
\vspace{-0.5em}
\noindent The sparse BFS trees reduce the effective distance from~68 to~30.  The ICI algorithm closes the remaining gap in a single cycle expansion round, producing a verified 32-operation path: $4+24+4$.
\end{example}

\subsection{Path post-processing}

The paths produced by ICI are generally not shortest.  The package provides two post-processing utilities:
\begin{itemize}
\item \texttt{short\_position}: algebraic simplification that cancels inverse pairs ($LR\to\varepsilon$), reduces consecutive shifts modulo~$n$, and simplifies blocks between reversals.
\item \texttt{short\_path\_bfs}: depth-limited BFS hopping.  At each position along the path, a local BFS of depth~$d$ explores nearby states; if any of them appears later in the original path, the algorithm jumps ahead, replacing the intervening segment with the shorter BFS route.
\end{itemize}

\subsection{C++ state store}
\label{subsec:statestore}

A central data structure is the \texttt{StateStore}, implemented entirely in C++.  It provides compact, incremental, hash-indexed storage of permutation states and is designed for the accumulate-and-intersect pattern of Algorithm~\ref{alg:ici}.

Each entry in the store consists of a permutation vector of fixed length~$n$ together with metadata: step index within the combination, combination index, cycle number, generator label, and celestial coordinates.  States are stored contiguously in a flat array; an open-addressing hash table maps each unique permutation (keyed by a combined hash of its elements) to the list of array indices where it occurs.  This design achieves $O(1)$~amortized insertion and lookup.

The key operations are:
\begin{itemize}
\item \texttt{store\_add\_from\_df} and \texttt{store\_analyze\_combos}: bulk insertion from R data frames or direct C++ cycle expansion.
\item \texttt{store\_find\_intersections}: given two stores $\mathcal{S}$ and $\mathcal{F}$, returns the set of state keys present in both, in $O(\min(|\mathcal{S}|,|\mathcal{F}|))$ time.
\item \texttt{store\_reconstruct\_path}: traces the chain of bridge states backward through successive cycles to assemble the full operation sequence.
\item \texttt{store\_filter\_middle} and \texttt{store\_set\_opd}: filtering operations that restrict queries to interior states of cycles or specific combination subsets, reducing noise in bridge selection.
\end{itemize}

\subsection{Celestial coordinates}
\label{subsec:celestial}

Each state visited during cycle expansion is annotated with \emph{celestial coordinates}, providing a geometric embedding of the search trajectory.  The construction is inspired by the celestial holography program of Pasterski~\cite{pasterski2021}, which establishes a duality between scattering amplitudes in four-dimensional flat spacetime and correlators of a two-dimensional conformal field theory on the celestial sphere.  We adapt this idea to the discrete setting of Cayley graphs: cumulative operation counts are mapped to points on~$S^2$ via stereographic projection.

Let $(n_L,n_R,n_X)$ denote the cumulative counts of left shifts, right shifts, and reversals applied along the path from the seed state.  Define the complex coordinate
\[
  z = \frac{n_L - n_R}{n_L + n_R + n_X + 1} + i\,\frac{n_X}{n_L + n_R + n_X + 1}.
\]
The $+1$ in the denominator ensures $z$ is well-defined at the seed state itself, where $(n_L,n_R,n_X)=(0,0,0)$; in this case $z=0$, corresponding to the origin of the stereographic plane.
Apply the inverse stereographic projection to obtain a point on~$S^2$, yielding spherical coordinates $(\theta,\phi)$ and a conformal parameter~$\omega$.  In the current implementation, these coordinates serve primarily as a \emph{visualization tool}, mapping the search history onto the sphere to reveal geometric structure in the exploration pattern (see Figure~\ref{fig:vis2}).  An angular distance $d_z(\sigma,\tau)$ between celestial positions is also available and can in principle be used for bridge selection; however, experiments to date have not shown this to be more effective than Manhattan or breakpoint distance on the permutation vectors themselves.  We regard celestial coordinates as an experimental feature whose full potential---particularly in connection with the holographic framework discussed in Section~\ref{sec:discussion}---remains to be explored.

\subsection{GPU acceleration}
\label{subsec:gpu}

When the \texttt{ggmlR} package is installed and a Vulkan-capable GPU is detected, \texttt{cayleyR} offloads two computational bottlenecks to the GPU.

\subsubsection{Batch state transformation}
Each generator $L$, $R$, $X_k$ can be represented as an $n\times n$ permutation matrix.  A composed operation sequence~$w$ corresponds to the matrix product $M_w = M_{w_m}\cdots M_{w_1}$.  Applying $w$ to a batch of $B$~states is then a single matrix multiplication $M_w \cdot S$, where $S$ is the $n\times B$ state matrix.  The function \texttt{apply\_operations\_batch\_gpu} performs this multiplication on the GPU, and \texttt{store\_analyze\_combos\_gpu} integrates it directly into the cycle expansion loop, bypassing the R layer entirely.

\subsubsection{Pairwise distance computation}
Computing Manhattan distances between all pairs of states in two sets of sizes $N_1$ and $N_2$ is an $O(N_1 N_2 n)$ operation that dominates the bridge selection phase for large stores.  The function \texttt{manhattan\_distance\_matrix\_gpu} performs this computation on the GPU in batches, avoiding host-memory overflow for large state sets.

\section{Computational Experiments}
\label{sec:experiments}

We present preliminary benchmarks illustrating the behavior of the ICI algorithm under different parameter settings.  All experiments were conducted using \texttt{cayleyR} version~0.1.0 on a Linux workstation with an AMD Ryzen~5 5600 (6~cores, 12~threads).  Note: the current CRAN release is version~0.2.1; the core ICI algorithm and its parameters are unchanged between versions, so the results remain representative.

\subsection{Effect of the ranking criterion}

The \texttt{sort\_by} parameter controls which random operation sequences are retained after the sampling phase (line~7 of Algorithm~\ref{alg:ici}).  We compare six strategies across 12~random TopSpin instances with $n$ ranging from~14 to~30, $k=4$, \texttt{n\_moves}${}=100$, \texttt{max\_iterations}${}=200$, and a 30-second timeout per strategy.  The start state is the identity and the target is a random permutation.

\begin{table}[ht]
\centering
\caption{Aggregated \texttt{sort\_by} benchmark: success rate and median time across 12~instances ($n=14$--$30$, $k=4$).}
\label{tab:sortby}
\small
\begin{tabular}{lccc}
\hline
\texttt{sort\_by} & Success rate & Median time (s) & Typical path length \\
\hline
\texttt{longest}              & 7/12 (58\%) & 0.4--60  & 4--7644 \\
\texttt{most\_unique}         & 10/12 (83\%) & 0.2--38  & 6--73980 \\
\texttt{most\_repeated}       & 9/12 (75\%) & 0.0--7.8 & 4--2042 \\
\texttt{least\_repeated}      & 3/12 (25\%) & 0.0--28  & 4--8 \\
\texttt{longest+most\_unique} & 7/12 (58\%) & 0.2--161 & 4--37269 \\
\texttt{most\_unique+longest} & 7/12 (58\%) & 0.2--161 & 4--34146 \\
\hline
\end{tabular}
\end{table}

Table~\ref{tab:sortby} summarizes the results.  Several patterns emerge from the 12~runs.

\texttt{most\_unique} achieves the highest success rate (83\%), finding paths in 10 out of 12~instances.  However, the paths it produces are often very long (up to 73\,980~operations), since it prioritizes state diversity over cycle structure.

\texttt{most\_repeated} is the most reliable fast strategy: 75\% success rate with consistently short paths (typically under 2000~operations) and very low search times (median under 1~second).  It selects sequences whose cycles revisit many states, which concentrates the search and increases intersection probability in dense regions of the graph.

\texttt{longest} succeeds in 58\% of cases.  When it works, it often finds short paths quickly (e.g., 4--6~operations in easy instances), but it frequently times out on harder instances where long cycles fail to overlap.

\texttt{least\_repeated} is the weakest strategy (25\% success), confirming that sequences with few repeated states produce sparse, non-overlapping cycles.

The combined strategies (\texttt{longest+most\_unique} and \texttt{most\_unique+longest}) do not consistently outperform the single-criterion strategies.  In some instances they find optimal short paths; in others they timeout.  Their behavior is sensitive to the interaction between the two ranking stages.

A key observation is that no single strategy dominates across all instances.  The optimal choice depends on the specific permutation and puzzle size: \texttt{most\_repeated} is a safe default for reliability, while \texttt{most\_unique} maximizes coverage at the cost of path length.  This variability motivates future work on adaptive strategy selection.

\subsection{Scalability with scramble distance}
\label{subsec:scramble}

To assess how the algorithm scales with the difficulty of the target state, we fix
$n=14$, $k=4$ and vary the \emph{scramble distance} \texttt{n\_moves} from~20 to~150
(the number of random moves used to generate the target from the identity). The algorithm
is \texttt{find\_path\_bfs} with \texttt{bfs\_levels}${}=200$, \texttt{bfs\_n\_hubs}${}=7$,
\texttt{bfs\_n\_random}${}=3$,
\texttt{sort\_by}${}=$\texttt{c("longest",\,"most\_unique")},
\texttt{combo\_length}${}=25$, \texttt{n\_samples}${}=400$, \texttt{n\_top}${}=100$,
\texttt{max\_iterations}${}=150$, \texttt{potc}${}=1$, \texttt{ptr}${}=3$,
\texttt{opd}${}=$TRUE, \texttt{distance\_method}${}=$\texttt{"manhattan"}.
Path shortening is applied via \texttt{short\_path\_bfs}. The column \emph{Cycles}
reports the number of ICI rounds executed in Step~4 (hub-to-hub search) of
\texttt{find\_path\_bfs}.

\begin{table}[ht]
\centering
\caption{ICI performance on TopSpin$(14,4)$ as a function of scramble distance.}
\label{tab:nmoves}
\small
\begin{tabular}{rrrrrrrr}
\hline
\texttt{n\_moves} & Cycles & Path & Short & Savings & Search (s) & Short (s) & Total (s) \\
\hline
  20 &  0 &     6 &     6 &    0 & 0.67 & 0.00 & 0.67 \\
  30 &  1 &    68 &     8 &   60 & 1.64 & 0.00 & 1.64 \\
  40 &  5 &   740 &   730 &   10 & 0.63 & 1.08 & 1.71 \\
  50 & 11 &  2461 &  1689 &  772 & 1.78 & 2.50 & 4.28 \\
  60 &  5 &   718 &   718 &    0 & 0.66 & 1.08 & 1.74 \\
  70 &  5 &  1281 &   943 &  338 & 0.71 & 1.41 & 2.12 \\
  80 & 12 &  1942 &  1316 &  626 & 3.47 & 2.02 & 5.49 \\
  90 &  3 &   430 &   366 &   64 & 0.37 & 0.53 & 0.90 \\
 100 &  3 &   477 &   467 &   10 & 0.46 & 0.70 & 1.16 \\
 110 &  5 &   793 &   767 &   26 & 0.76 & 1.14 & 1.90 \\
 120 &  5 &  1204 &   607 &  597 & 0.67 & 0.91 & 1.57 \\
 130 &  7 &  1777 &  1114 &  663 & 1.13 & 1.65 & 2.78 \\
 140 &  7 &  1342 &   643 &  699 & 1.09 & 0.94 & 2.04 \\
 150 & 12 &  2102 &  1443 &  659 & 1.71 & 2.16 & 3.86 \\
\hline
\end{tabular}
\end{table}

All 14~instances were solved successfully (Table~\ref{tab:nmoves}). Several observations
stand out. First, the search time remains modest across all scramble distances: the mean
search time is 1.12~seconds, and even the hardest instance ($\texttt{n\_moves}=80$,
requiring 12~cycles) is solved in 3.5~seconds. The total time including path shortening
averages 2.3~seconds.

Second, the number of ICI cycles required is not monotonically related to the scramble
distance. Some instances at $\texttt{n\_moves}=90$ or~100 are solved in 3~cycles, while
$\texttt{n\_moves}=50$ requires~11. This reflects the stochastic nature of the algorithm:
the difficulty depends not just on the graph distance but on the particular cycle structure
encountered during random sampling.

Third, the path shortening step (\texttt{short\_path\_bfs}) provides substantial
compression in many cases, reducing path length by up to 50\% (e.g., from 1342 to~643
at $\texttt{n\_moves}=140$). However, the savings are inconsistent: at
$\texttt{n\_moves}=60$ the shortener achieves zero savings, while at
$\texttt{n\_moves}=50$ it removes 772~operations. The shortening time is comparable
to the search time, roughly doubling the total computation.

Experiments with GPU acceleration and scalability for $n$ exceeding~20 will be reported
in a future revision.

\subsection{Scalability with puzzle size}

We also test how the algorithm scales with the puzzle size~$n$, fixing $k=4$ and $\texttt{n\_moves}=20$ while varying $n$ from~10 to~16.  The algorithm and parameters are identical to Section~\ref{subsec:scramble}: \texttt{find\_path\_bfs} with \texttt{bfs\_levels}${}=200$, \texttt{combo\_length}${}=25$, \texttt{n\_samples}${}=400$, \texttt{n\_top}${}=100$, \texttt{sort\_by}${}=$\texttt{c("longest",\,"most\_unique")}.

\begin{table}[ht]
\centering
\caption{ICI performance as a function of puzzle size~$n$ ($k=4$, $\texttt{n\_moves}=20$).}
\label{tab:nsize}
\small
\begin{tabular}{rrrrrrrrr}
\hline
$n$ & Cycles & Path & Short & Savings & Search (s) & Short (s) & Total (s) \\
\hline
10 & 1 &  372 &  163 & 209 & 0.25 & 0.21 & 0.46 \\
11 & 1 &  244 &    8 & 236 & 0.20 & 0.00 & 0.20 \\
12 & 2 & 1138 & 1138 &   0 & 0.42 & 1.66 & 2.08 \\
13 & 2 &  296 &    8 & 288 & 0.47 & 0.00 & 0.47 \\
14 & 1 &  144 &    8 & 136 & 0.22 & 0.00 & 0.22 \\
15 & 1 &    2 &    2 &   0 & 0.32 & 0.00 & 0.32 \\
16 & 0 &    2 &    2 &   0 & 0.12 & 0.00 & 0.12 \\
\hline
\end{tabular}
\end{table}

All seven instances were solved (Table~\ref{tab:nsize}).  The search time remains under 0.5~seconds for all sizes, with no visible growth trend from $n=10$ to $n=16$.  This is consistent with the design of the ICI algorithm: the per-cycle cost is dominated by the number of states generated (controlled by \texttt{n\_samples} and \texttt{combo\_length}), not by $n!$.  The state space grows super-exponentially with~$n$, but the algorithm only ever materializes a small fraction of it.

The path lengths and cycle counts vary considerably across instances, again reflecting the stochastic nature of the method.  Notably, the $n=15$ and $n=16$ instances are solved with paths of only 2~operations---the random target happened to be very close to the identity---while $n=12$ requires 2~cycles and produces a long path (1138) that the shortener fails to compress.  These extremes illustrate that the algorithm's behavior is primarily governed by the intrinsic difficulty of the particular instance rather than the nominal size of the state space.

\section{Discussion and Future Directions}
\label{sec:discussion}

\subsection{Toward holographic cycle intersection}

A speculative but promising direction comes from the recently proposed holographic duality framework for Cayley graphs \cite{chervov2025holo}.  In that work, vertices of the Cayley graph $\Gamma(S_n,S)$ are mapped to lattice paths inside a planar polygon, with the key property that word metrics (distances in the graph) correspond to areas under the respective paths---a discrete realization of the ``complexity $=$ volume'' principle from AdS/CFT holography.

This geometric perspective suggests an alternative approach to cycle intersection detection.  A cycle in the Cayley graph---a closed walk generated by iterating an operation sequence---would map to a closed lattice path in the polygon.  Two cycles emanating from different seed states would map to two families of paths, and their intersection in the graph would correspond to a geometric intersection of paths in two dimensions.  Finding such intersections among piecewise-linear curves in~$\mathbb{R}^2$ is in principle far cheaper than hash-based comparison in the full $n!$-element state space.

\begin{figure}[ht]
\centering
\includegraphics[width=0.55\textwidth]{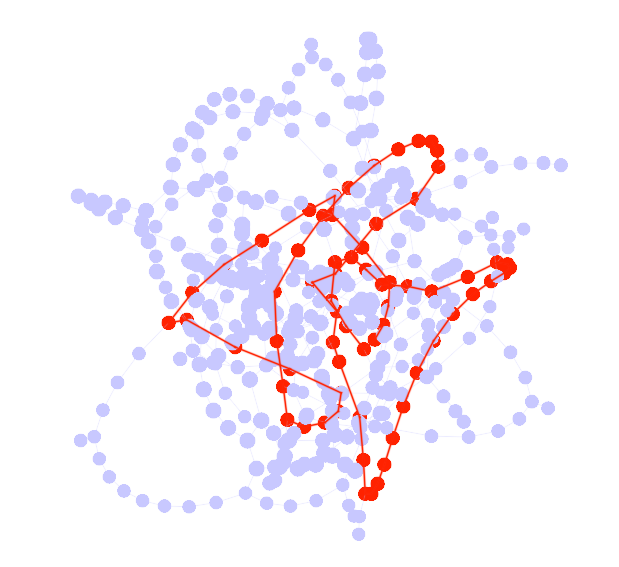}
\caption{Three-dimensional layout of cycle states for TopSpin$(7,4)$ using celestial coordinates $(\theta,\phi,\omega)$.  Light blue points represent all states visited during cycle analysis of multiple operation sequences; edges connect consecutive states within each cycle.  The red overlay highlights the longest cycle found.  Visualization produced by the \texttt{cgvR} package (in development).}
\label{fig:vis1}
\end{figure}

\begin{figure}[ht]
\centering
\includegraphics[width=0.55\textwidth]{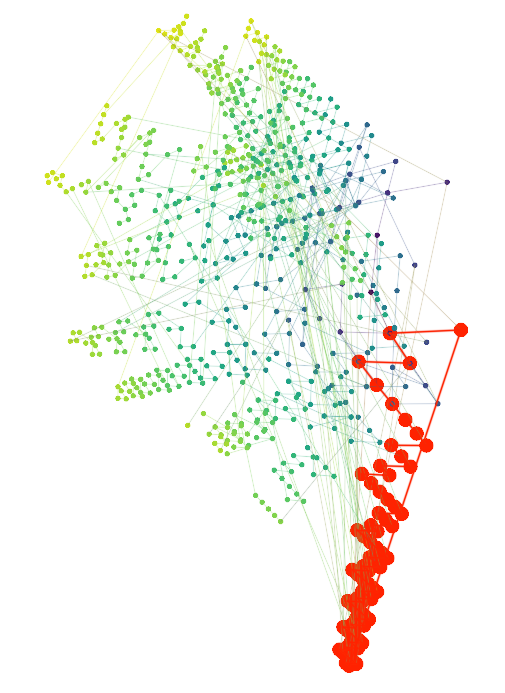}
\caption{Celestial coordinate embedding of cycle expansion states for TopSpin$(20,4)$.  Each point is a permutation state mapped to~$S^2$ via stereographic projection of cumulative operation counts $(n_L,n_R,n_X)$; colors encode different operation sequences.  The red chain traces the recovered path through successive bridge states.  Spatial clustering reflects that similar operation sequences produce nearby celestial positions.}
\label{fig:vis2}
\end{figure}

The \texttt{cayleyR} package already implements a coordinate embedding (celestial coordinates, Section~\ref{subsec:celestial}) that maps operation counts to points on the unit sphere.  While this embedding is useful for visualization, it does not yet capture the algebraic structure needed for reliable intersection detection---in practice, states that are close on the sphere are not necessarily close in the Cayley graph.  A forthcoming visualization module will provide tools for exploring these geometric representations; sample outputs are shown in Figures~\ref{fig:vis1}--\ref{fig:vis2}.

Significant challenges remain.  The holographic map in \cite{chervov2025holo} is established for Cayley graphs generated by transpositions and specific generator sets; extending it to the TopSpin generators $\{L, R, X_k\}$ requires further theoretical work.  The lift from polygon coordinates back to permutations must be computationally efficient, and it is not yet clear whether the map preserves enough structure for the TopSpin case to make geometric intersection detection practical.  We regard this as an open problem for future investigation.

\subsection{Further directions}

The current implementation of \texttt{cayleyR} is focused exclusively on the TopSpin$(n,k)$ puzzle: the generators $\{L,R,X_k\}$, the cycle analysis routines, and the path post-processing are all tailored to the shift-and-reverse structure of TopSpin.  However, the underlying algorithmic ideas---iterative cycle intersection, hash-indexed state stores, bridge selection by distance heuristics---are not inherently tied to this particular puzzle.  In principle, any permutation puzzle whose moves can be expressed as generators of a subgroup of~$S_n$ could be attacked with the same framework.  Natural candidates include pancake sorting (prefix reversals of variable length), burnt pancake (signed prefix reversals), and Hungarian Rings.

Several concrete extensions are planned for future releases:
\begin{itemize}
\item \emph{Coordinate-guided search.}  Using celestial or holographic coordinates to steer the cycle expansion toward the target, rather than relying solely on distance heuristics applied to permutation vectors.  This includes directional search: biasing the generation of operation sequences toward directions in coordinate space that point from the current state toward the target.
\item \emph{Multi-point pathfinding.}  Accepting a list of waypoints and finding a path that visits them sequentially, reusing the accumulated state store between successive searches.  This extends the single-pair ICI to a traveling-salesman-like regime where the search space built for one pair informs the next.
\item \emph{Combination library.}  Pre-generating and caching libraries of high-quality operation sequences (ranked by cycle length or uniqueness) that can be reused across multiple pathfinding queries, amortizing the sampling cost.
\item \emph{Visualization package.}  A companion R package (\texttt{cgvR}, currently under development) will provide interactive visualization of cycles, state stores, bridge trajectories, and coordinate embeddings on the celestial sphere.
\item \emph{Generalization beyond TopSpin.}  A modular generator interface allowing users to define custom move sets, with adapted path inversion and simplification routines for arbitrary generator--inverse pairs.
\end{itemize}

Other directions under consideration include adaptive combination selection (using reinforcement learning to bias toward sequences whose cycles are more likely to intersect), and parallel search (multiple workers expanding cycles from different bridge states with a shared state store).

\section{Conclusion}
\label{sec:conclusion}

\textit{}

We have presented \texttt{cayleyR}, an R package that solves the TopSpin$(n,k)$ puzzle by detecting cycle intersections in Cayley graphs of the symmetric group.  The core contribution is the Iterative Cycle Intersection (ICI) algorithm, which generates random operation sequences from both the start and target states, expands their cycles, and searches for shared permutation states via hash-indexed lookup.  When no intersection is found, distance-guided bridge selection narrows the gap and the process repeats.

The algorithm is complemented by a hub-based transport network strategy that builds sparse BFS trees from both endpoints, reducing the effective search distance before invoking ICI.  The implementation combines R with a C++ backend (via Rcpp) featuring an open-addressing hash-indexed state store, OpenMP-parallelized combination evaluation, and optional Vulkan GPU acceleration for batch state transformations and pairwise distance computation.

Computational experiments on TopSpin$(14,4)$ and TopSpin$(10{-}16,4)$ demonstrate that the algorithm reliably finds paths in under a few seconds, with the ranking criterion for operation sequences having a significant impact on path quality.  The celestial coordinate embedding provides a visualization framework for the search geometry, and a companion visualization package (\texttt{cgvR}) is under development.

The current implementation is specific to the TopSpin puzzle, but the underlying ideas---cycle intersection, bridge selection, hash-indexed state stores---are applicable to any permutation puzzle expressible as a Cayley graph.  Future directions include coordinate-guided search, multi-point pathfinding, and integration with the holographic duality framework recently proposed for Cayley graphs of~$S_n$.

The package is open-source and available on CRAN \cite{cayleyR}. A companion visualization package, \texttt{cgvR}, is also available on CRAN \cite{cgvR}.

\end{document}